\documentclass{article} 
\def\arxiv{0}
\usepackage{amsmath,mathtools,amssymb,amsthm}
\if\arxiv1
\usepackage{times}
\usepackage[left=1in,right=1in,top=1in,bottom=1in]{geometry}
\else
\usepackage{iclr2016_conference, times}
\iclrfinalcopy
\fi

\newcommand{\bg}{\mathbf{g}}

\newcommand{\br}{\mathbf{r}}

\usepackage{bm}

\newcommand{\Real}{\mathbb{R}}

\DeclareMathOperator*{\minimize}{minimize}

\newcommand{\defeq}{\vcentcolon=}
\newcommand{\given}[1][]{\:#1\vert\:}

\newcommand{\grad}{\nabla}

\newcommand{\E}{\mathbb{E}}

\newcommand{\Ea}[1]{\E\left[#1\right]}
\newcommand{\Eb}[2]{\E_{#1}\left[#2\right]}

\DeclarePairedDelimiter{\norm}{\|}{\|}

\DeclarePairedDelimiter{\lrbrace}{ \{ }{ \} }
\DeclarePairedDelimiter{\lrparen}{(}{)}

\newcommand{\Qpi}{\ensuremath{Q^{\pi}}}
\newcommand{\Api}{\ensuremath{A^{\pi}}}
\newcommand{\Vpi}{\ensuremath{V^{\pi}}}

\newcommand{\thold}{\theta_{old}}

\newcommand{\pith}{\pi_{\theta}}

\newcommand{\gradth}{\grad_{\theta}}

\newcommand{\kl}[2]{D_{KL}(#1 \ \| \ #2)}

\newcommand{\tdlam}{TD($\lambda$)}

\newcommand{\tdone}{TD($1$)}

\usepackage{graphicx} 
\usepackage{subfigure}
\usepackage{aliascnt}
\usepackage{algorithm}
\usepackage{algpseudocode}
\usepackage{multicol}
\usepackage{enumitem}
\usepackage{units}

\usepackage{hyperref}
\hypersetup{
colorlinks   = true, 
urlcolor     = blue, 
linkcolor    = blue, 
citecolor   = black 
}
\usepackage{color}

\newtheorem{defn}{Definition}
\newtheorem{prop}{Proposition}

\usepackage[capitalize]{cleveref}

\renewcommand{\bg}{g}

\newcommand{\dv}{\delta^V}

\newcommand{\hata}{\hat{A}}

\newcommand{\hatalam}{\hat{A}^{\mathrm{GAE(\gamma,\lambda)}}}

\newcommand{\phold}{\phi_{\text{old}}}
\newcommand{\skipforicml}[1]{}
\newcommand{\delay}{l}
\newcommand{\meankl}[1]{{\ensuremath \overline D_{\rm KL}^{#1}}}
\newcommand{\bgrad}{\bg^{\gamma}}
\newcommand{\Apigam}{A^{\pi,\gamma}}
\newcommand{\Qpigam}{Q^{\pi,\gamma}}
\newcommand{\Vpigam}{V^{\pi,\gamma}}

\newcommand{\tilr}{\tilde{r}}
\newcommand{\resp}{\chi}
\newcommand{\Vhat}{\hat{V}}
\newcommand{\bofpast}{b_t(s_{0:t},a_{0:t-1})}
\newcommand{\qofall}{Q_t(s_{0:\infty},a_{0:\infty})}
\newcommand{\just}{$\gamma$-just}
\newcommand{\GAE}{\mathrm{GAE}}

\author{John~Schulman, Philipp~Moritz, Sergey~Levine, Michael~I.~Jordan and Pieter~Abbeel \\
Department of Electrical Engineering and Computer Science\\
University of California, Berkeley\\
\texttt{\{joschu,pcmoritz,levine,jordan,pabbeel\}@eecs.berkeley.edu}
}
\title{High-Dimensional Continuous Control Using Generalized Advantage Estimation}

\begin{document}

\if\arxiv1
\vspace{1cm}
\begin{center}
{\LARGE {\bf{High-Dimensional Continuous Control \\ \vspace{.15cm} Using Generalized Advantage Estimation}}} \\
\vspace{.5cm}
{\large
John~Schulman ~~~~~~
Philipp~Moritz ~~~~~~
Sergey~Levine ~~~~~~ \\
\vspace{.1cm}
Michael~Jordan ~~~~~~
Pieter~Abbeel
} \\
\vspace{.2cm}
{\tt \{joschu,pcmoritz,svlevine,jordan,pabbeel\}@eecs.berkeley.edu} \\
\vspace{.2cm}
{\large
Department of Electrical Engineering and Computer Science \\
\vspace{.1cm}
University of California, Berkeley
}
\vspace*{.2in}
\end{center}
\else
\maketitle
\fi

\begin{abstract}

Policy gradient methods are an appealing approach in reinforcement learning because they directly optimize the cumulative reward and can straightforwardly be used with nonlinear function approximators such as neural networks.
The two main challenges are the large number of samples typically required, and  the difficulty of obtaining stable and steady improvement despite the nonstationarity of the incoming data.
We address the first challenge by using value functions to
substantially reduce the variance of policy gradient estimates at the
cost of some bias, with an exponentially-weighted estimator of the
advantage function that is analogous to \tdlam{}.
We address the second challenge by using trust region optimization procedure for both the policy and the value function, which are represented by neural networks.

Our approach yields strong empirical results on highly challenging 3D
locomotion tasks, learning running gaits for bipedal and quadrupedal
simulated robots, and learning a policy for getting the biped to stand up from starting out lying on the ground.
In contrast to a body of prior work that uses hand-crafted policy
representations, our neural network policies map directly from raw
kinematics to joint torques.
Our algorithm is fully model-free, and the amount of simulated experience required for the learning tasks on 3D bipeds corresponds to 1-2 weeks of real time.

\end{abstract}

\section{Introduction}

The typical problem formulation in reinforcement learning is to maximize the expected total reward of a policy.
A key source of difficulty is the long time delay between actions and their positive or negative effect on rewards;
this issue is called the \textit{credit assignment problem} in the reinforcement learning literature \citep{minsky1961steps,sutton1998introduction}, and the \textit{distal reward problem} in the behavioral literature \citep{hull1943principles}.
Value functions offer an elegant solution to the credit assignment problem---they allow us to estimate the goodness of an action before the delayed reward arrives.
Reinforcement learning algorithms make use of value functions in a variety of different ways; this paper considers algorithms that optimize a parameterized policy and use value functions to help estimate how the policy should be improved.

When using a parameterized \textit{stochastic} policy, it is possible to obtain an unbiased estimate of the gradient of the expected total returns \citep{williams1992simple,sutton1999policy,baxter2000reinforcement}; these noisy gradient estimates can be used in a stochastic gradient ascent algorithm.
Unfortunately, the variance of the gradient estimator scales unfavorably with the time horizon, since the effect of an action is confounded with the effects of past and future actions.
Another class of policy gradient algorithms, called actor-critic methods, use a value function rather than the empirical returns, obtaining an estimator with lower variance at the cost of introducing bias \citep{konda2003onactor,hafner2011reinforcement}.
But while high variance necessitates using more samples,
bias is more pernicious---even with an unlimited number of samples, bias can cause the algorithm to fail to converge, or to converge to a poor solution that is not even a local optimum.

We propose a family of policy gradient estimators that significantly reduce variance while maintaining a tolerable level of bias.
We call this estimation scheme, parameterized by $\gamma \in [0,1]$ and $\lambda \in [0,1]$, the generalized advantage estimator (GAE).
Related methods have been proposed in the context of online actor-critic methods \citep{kimura1998analysis,wawrzynski2009real}. We provide a more general analysis, which is applicable in both the online and batch settings, and discuss an interpretation of our method as an instance of reward shaping \citep{ng1999policy}, where the approximate value function is used to shape the reward.

We present experimental results on a number of highly challenging 3D locomotion tasks, where we show that our approach can learn complex gaits using high-dimensional, general purpose neural network function approximators for both the policy and the value function, each with over $10^4$ parameters.
The policies perform torque-level control of simulated 3D robots with up to 33 state dimensions and 10 actuators.

The contributions of this paper are summarized as follows:
\begin{enumerate}[leftmargin=*]
\item We provide justification and intuition for an effective variance reduction scheme for policy gradients, which we call generalized advantage estimation (GAE). While the formula has been proposed in prior work \citep{kimura1998analysis,wawrzynski2009real}, our analysis is novel and enables GAE to be applied with a more general set of algorithms, including the batch trust-region algorithm we use for our experiments.
\item We propose the use of a trust region optimization method for the value function, which we find is a robust and efficient way to train neural network value functions with thousands of parameters.
\item By combining (1) and (2) above, we obtain an algorithm that empirically is effective at learning neural network policies for challenging control tasks. The results extend the state of the art in using reinforcement learning for high-dimensional continuous control. Videos are available at \url{https://sites.google.com/site/gaepapersupp}.
\end{enumerate}

\section{Preliminaries}
\label{sec:prelim}

We consider an undiscounted formulation of the policy optimization problem.
The initial state  $s_0$ is sampled from distribution $\rho_0$.
A trajectory $(s_0, a_0, s_1, a_1, \dots)$ is generated by sampling actions according to the policy $a_t \sim \pi(a_t \given s_t)$ and sampling the states according to the dynamics $s_{t+1} \sim P(s_{t+1} \given s_t, a_t)$, until a terminal (absorbing) state is reached.
A reward $r_t = r(s_t,a_t,s_{t+1})$ is received at each timestep.
The goal is to maximize the expected total reward $\sum_{t=0}^{\infty} r_t$, which is assumed to be finite for all policies.
Note that we are not using a discount as part of the problem specification; it will appear below as an algorithm parameter that adjusts a bias-variance tradeoff.
But the discounted problem (maximizing $\sum_{t=0}^{\infty} \gamma^t r_t$) can be handled as an instance of the undiscounted problem in which we absorb the discount factor into the reward function, making it time-dependent.

Policy gradient methods maximize the expected total reward by repeatedly estimating the gradient $\bg \defeq \gradth \Ea{\sum_{t=0}^{\infty} r_t}$.
There are several different related expressions for the policy gradient, which have the form
\begin{align}\label{eq:pg-abstract}
\bg =
\Ea{\sum_{t=0}^{\infty} \Psi_t \gradth \log \pith(a_t \given s_t)},
\end{align}
where $\Psi_t$ may be one of the following:
\begin{multicols}{2}
\begin{enumerate}
\item $\sum_{t=0}^{\infty} r_{t}$: total reward of the trajectory. \label{eq:raw-reinforce}
\item $\sum_{t'=t}^{\infty} r_{t'}$: reward following action $a_t$. \label{eq:q-reinforce}
\item $\sum_{t'=t}^{\infty} r_{t'} - b(s_t)$: baselined version of previous formula. \label{eq:q-reinforce-wb}
\item $\Qpi(s_t, a_t)$: \label{eq:qfunc-reinforce} state-action value function.
\item $\Api(s_t, a_t)$: \label{eq:adv-reinforce} advantage function.
\item $r_{t} + \Vpi(s_{t+1}) - \Vpi(s_t)$: TD residual.\label{eq:ac-reinforce}
\end{enumerate}
\end{multicols}
The latter formulas use the definitions
\begin{align}
\Vpi(s_t) &\defeq \Eb{\substack{s_{t+1:\infty},\\a_{t:\infty} } }{\sum_{\delay=0}^{\infty} r_{t+\delay}} \hspace{0.5in}
\Qpi(s_t,a_t) \defeq \Eb{\substack{s_{t+1:\infty},\\a_{t+1:\infty}}}{\sum_{\delay=0}^{\infty} r_{t+\delay}} \\
\Api(s_t, a_t) &\defeq \Qpi(s_t, a_t) - \Vpi(s_t), \quad \text{(Advantage function)}.
\end{align}
Here, the subscript of $\mathbb{E}$ enumerates the variables being integrated over, where states and actions are sampled sequentially from the dynamics model $P(s_{t+1} \given s_t, a_t)$ and policy $\pi(a_t \given s_t)$, respectively.
The colon notation $a:b$ refers to the inclusive range $(a,a+1,\dots,b)$.
These formulas are well known and straightforward to obtain; they follow directly from Proposition \hyperlink{justprop}{1}, which will be stated shortly.

The choice $\Psi_t =\Api(s_t, a_t)$ yields almost the lowest possible variance, though in practice, the advantage function is not known and must be estimated.
This statement can be intuitively justified by the following interpretation of the policy gradient: that a step in the policy gradient direction should increase the probability of better-than-average actions and decrease the probability of worse-than-average actions.
The advantage function, by it's definition $\Api(s,a)=\Qpi(s,a)-\Vpi(s)$, measures whether or not the action is better or worse than the policy's default behavior.
Hence, we should choose $\Psi_t$ to be the advantage function $\Api(s_t, a_t)$, so that the gradient term $\Psi_t \gradth \log \pith(a_t \given s_t)$ points in the direction of increased $\pith(a_t \given s_t)$ if and only if $\Api(s_t, a_t)>0$.
See \cite{greensmith2004variance} for a more rigorous analysis of the variance of policy gradient estimators and the effect of using a baseline.

We will introduce a parameter $\gamma$ that allows us to reduce variance by downweighting rewards corresponding to delayed effects, at the cost of introducing bias.
This parameter corresponds to the discount factor used in discounted formulations of MDPs, but we treat it as a variance reduction parameter in an undiscounted problem; this technique was analyzed theoretically by \cite{marbach2003approximate,kakade2001optimizing,thomas2014bias}.
The discounted value functions are given by:
\begin{align}
\Vpigam(s_t) &\defeq \Eb{\substack{s_{t+1:\infty},\\a_{t:\infty} } }{\sum_{\delay=0}^{\infty} \gamma^{\delay} r_{t+\delay}}\hspace{0.5in}
\Qpigam(s_t,a_t) \defeq \Eb{\substack{s_{t+1:\infty},\\a_{t+1:\infty}}}{\sum_{\delay=0}^{\infty} \gamma^{\delay} r_{t+\delay}}\\
\Apigam(s_t, a_t)&\defeq\Qpigam(s_t,a_t) - \Vpigam(s_t).
\end{align}

The discounted approximation to the policy gradient is defined as follows:
\begin{align}
\bgrad
&\defeq \Eb{\substack{s_{0:\infty}\\ a_{0:\infty}}}{ \sum_{t=0}^{\infty}\Apigam(s_t,a_t) \gradth \log \pith(a_t \given s_t)}.
\label{eq:pg-adv-biased}
\end{align}

The following section discusses how to obtain biased (but not too biased) estimators for $\Apigam$, giving us noisy estimates of the discounted policy gradient in \Cref{eq:pg-adv-biased}.

Before proceeding, we will introduce the notion of a \just{} estimator of the advantage function, which is an estimator that does not introduce bias when we use it in place of $\Apigam$ (which is not known and must be estimated) in \Cref{eq:pg-adv-biased} to estimate $\bgrad$.\footnote{Note, that we have already introduced bias by using $\Apigam$ in place of $\Api$; here we are concerned with obtaining an unbiased estimate of $\bgrad$, which is a biased estimate of the policy gradient of the undiscounted MDP.}
Consider an advantage estimator $\hata_t(s_{0:\infty},a_{0:\infty})$, which may in general be a function of the entire trajectory.
\begin{defn}
The estimator $\hata_t$ is \just{} if
\begin{align}
\Eb{\substack{s_{0:\infty}\\ a_{0:\infty}}}{ \hata_t(s_{0:\infty},a_{0:\infty}) \gradth \log \pith(a_t \given s_t)}
=
\Eb{\substack{s_{0:\infty}\\ a_{0:\infty}}}{ \Apigam(s_t,a_t) \gradth \log \pith(a_t \given s_t)}.
\end{align}
\end{defn}
It follows immediately that if $\hata_t$ is \just{} for all $t$, then
\begin{align}
\Eb{\substack{s_{0:\infty}\\ a_{0:\infty}}}{ \sum_{t=0}^{\infty}\hata_t(s_{0:\infty},a_{0:\infty}) \gradth \log \pith(a_t \given s_t)}
=
\bgrad
\label{eq:ubgrad}
\end{align}

One sufficient condition for $\hata_t$ to be \just{} is that $\hata_t$ decomposes as the difference between two functions $Q_t$ and $b_t$, where $Q_t$ can depend on any trajectory variables but gives an unbiased estimator of the $\gamma$-discounted $Q$-function, and $b_t$ is an arbitrary function of the states and actions sampled before $a_t$.
\begin{prop} \hypertarget{justprop}{}
Suppose that $\hata_t$ can be written in the form $\hata_t(s_{0:\infty},a_{0:\infty}) = Q_t(s_{t:\infty},a_{t:\infty})-b_t(s_{0:t}, a_{0:t-1})$ such that for all $(s_t, a_t)$, $\Eb{s_{t+1:\infty},a_{t+1:\infty} \given s_t,a_t}{Q_t(s_{t:\infty},a_{t:\infty})}=\Qpigam(s_t, a_t)$. Then $\hata$ is \just{}.
\end{prop}

The proof is provided in \Cref{sec:proofs}.
It is easy to verify that the following expressions are \just{} advantage estimators for $\hata_t$:
\begin{multicols}{2}
\begin{itemize}
\item $\sum_{\delay=0}^{\infty} \gamma^{\delay} r_{t+\delay}$
\item $\Qpigam(s_t,a_t)$
\item $\Apigam(s_t,a_t)$
\item $r_t + \gamma \Vpigam(s_{t+1}) - \Vpigam(s_t)$.
\end{itemize}
\end{multicols}

\section{Advantage function estimation} \label{sec:advest}

\newcommand{\fut}{\substack{s_{t+1:\infty}\\ a_{t+1:\infty}}}

This section will be concerned with producing an accurate estimate $\hata_t$ of the discounted advantage function $\Apigam(s_t, a_t)$, which will then be used to construct a policy gradient estimator of the following form:
\begin{align}
\hat g = \frac{1}{N} \sum_{n=1}^N \sum_{t=0}^{\infty} \hata_t^n \gradth \log \pith(a_t^n \given s_t^n)
\label{eq:pg-gae}
\end{align}
where $n$ indexes over a batch of episodes.

Let $V$ be an approximate value function.
Define $\delta^V_t = r_t + \gamma V(s_{t+1}) - V(s_t)$, i.e., the TD residual of $V$ with discount $\gamma$ \citep{sutton1998introduction}.
Note that $\dv_t$ can be considered as an
estimate of the advantage of the action $a_t$. In fact, if we have the correct value function $V=\Vpigam$, then it is a \just{} advantage estimator, and in fact, an unbiased estimator of $\Apigam$:
\begin{align}
\Eb{s_{t+1}}{\delta^{\Vpigam}_t} &= \Eb{s_{t+1}}{r_t + \gamma \Vpigam(s_{t+1}) - \Vpigam(s_t)} \nonumber \\
&= \Eb{s_{t+1}}{\Qpigam(s_t, a_t) - \Vpigam(s_t)}= \Apigam(s_t, a_t).
\end{align}
However, this estimator is only \just{} for $V = \Vpigam$, otherwise it will yield biased policy gradient estimates.

Next, let us consider taking the sum of $k$ of these $\delta$ terms, which we will denote by $\hata_t^{(k)}$
\begin{alignat}{2}
\hata_t^{(1)} &\defeq  \dv_{t} &&= -V(s_t) + r_t + \gamma V(s_{t+1})\\
\hata_t^{(2)} &\defeq \dv_t + \gamma \dv_{t+1} &&= -V(s_t) + r_t + \gamma r_{t+1} + \gamma^2 V(s_{t+2}) \label{a2}\\
\hata_t^{(3)} &\defeq \dv_{t} + \gamma \dv_{t+1} + \gamma^2 \dv_{t+2} &&= -V(s_t) + r_t + \gamma r_{t+1} + \gamma^2 r_{t+2} + \gamma^3 V(s_{t+3}) \label{a3}
\end{alignat}
\begin{align}
\hata_t^{(k)} &\defeq \sum_{\delay=0}^{k-1} \gamma^{\delay} \dv_{t+l} = -V(s_t) + r_t + \gamma r_{t+1} + \dots + \gamma^{k-1} r_{t+k-1} + \gamma^{k} V(s_{t+k})
\end{align}
These equations result from a telescoping sum, and we see that $\hata_t^{(k)}$ involves a $k$-step estimate of the returns, minus a baseline term $-V(s_t)$.
Analogously to the case of $\dv_t = \hata_t^{(1)}$, we can consider $\hata_t^{(k)}$ to be an estimator of the advantage function, which is only \just{} when $V = \Vpigam$. However, note that the bias generally becomes smaller as $k \rightarrow \infty$, since the term $\gamma^k V(s_{t+k})$ becomes more heavily discounted, and the term $-V(s_t)$ does not affect the bias.
Taking $k \rightarrow \infty$, we get
\begin{align}
\hata_t^{(\infty)} = \sum_{\delay=0}^{\infty} \gamma^{\delay} \dv_{t+l} = -V(s_t) + \sum_{\delay=0}^{\infty} \gamma^{\delay} r_{t+\delay},
\end{align}
which is simply the empirical returns minus the value function baseline.

The generalized advantage estimator $\mathrm{GAE(\gamma,\lambda)}$ is defined as the exponentially-weighted average of these $k$-step estimators:
\begin{align}
\hatalam_t
&\defeq (1-\lambda)\lrparen*{ \hata_t^{(1)} + \lambda \hata_t^{(2)} + \lambda^2 \hata_t^{(3)} + \dots  }\nonumber \\
&= (1-\lambda)\lrparen*{ \dv_t + \lambda (\dv_t + \gamma \dv_{t+1}) + \lambda^2 (\dv_t + \gamma \dv_{t+1} + \gamma^2 \dv_{t+2}) + \dots }\nonumber \\
&= (1-\lambda)(
\dv_t (1 + \lambda + \lambda^2 + \dots)
+\gamma \dv_{t+1} (\lambda + \lambda^2 + \lambda^3 + \dots)\nonumber \\
&\ \ \ \ \  \ \ \ \ \ \ +\gamma^2 \dv_{t+2} (\lambda^2 + \lambda^3 + \lambda^4 + \dots)
+\dots)
\nonumber \\
&= (1-\lambda)\lrparen*{
\dv_t \lrparen*{\frac{1}{1-\lambda}}
+\gamma \dv_{t+1} \lrparen*{\frac{\lambda}{1-\lambda}}
+\gamma^2 \dv_{t+2} \lrparen*{\frac{\lambda^2}{1-\lambda}}
+\dots}
\nonumber \\
&= \sum_{\delay=0}^{\infty} (\gamma \lambda)^{\delay} \dv_{t+\delay}
\label{eq:gaelam1}
\end{align}
From \Cref{eq:gaelam1}, we see that the advantage estimator has a remarkably simple formula involving a discounted sum of Bellman residual terms.
\Cref{sec:shaping} discusses an interpretation of this formula as the returns in an MDP with a modified reward function.
The construction we used above is closely analogous to the one used to define \tdlam{} \citep{sutton1998introduction}, however \tdlam{} is an estimator of the value function, whereas here we are estimating the advantage function.

There are two notable special cases of this formula, obtained by setting $\lambda=0$ and $\lambda=1$.
\begin{alignat}{2}
\GAE(\gamma, 0):\ \ \ \hata_t &\defeq \delta_t &&= r_t + \gamma V(s_{t+1}) - V(s_t)\label{eq:adv-ac}\\
\GAE(\gamma, 1):\ \ \ \hata_t &\defeq \sum_{\delay=0}^{\infty} \gamma^{\delay} \delta_{t+\delay} &&= \sum_{\delay=0}^{\infty}\gamma^{\delay} r_{t+\delay} - V(s_t) \label{eq:adv-mc}
\end{alignat}
$\GAE(\gamma, 1)$ is \just{} regardless of the accuracy of $V$, but it has high variance due to the sum of terms.
$\GAE(\gamma, 0)$ is \just{} for $V = \Vpigam$ and otherwise induces bias, but it typically has much lower variance.
The generalized advantage estimator for $0 < \lambda < 1$ makes a compromise between bias and variance, controlled by parameter $\lambda$.

We've described an advantage estimator with two separate parameters $\gamma$ and $\lambda$, both of which contribute to the bias-variance tradeoff when using an approximate value function.
However, they serve different purposes and work best with different ranges of values.
$\gamma$ most importantly determines the scale of the value function $\Vpigam$, which does not depend on $\lambda$.
Taking $\gamma < 1$ introduces bias into the policy gradient estimate, regardless of the value function's accuracy.
On the other hand, $\lambda < 1$ introduces bias only when the value function is inaccurate.
Empirically, we find that the best value of $\lambda$ is much lower than the best value of $\gamma$, likely because $\lambda$ introduces far less bias than $\gamma$ for a reasonably accurate value function.

Using the generalized advantage estimator, we can construct a biased estimator of $\bgrad$, the discounted policy gradient from \Cref{eq:pg-adv-biased}:
\begin{align}
\bgrad &\approx \Ea{\sum_{t=0}^{\infty} \gradth \log \pith(a_t \given s_t) \hatalam_t}
= \Ea{\sum_{t=0}^{\infty} \gradth \log \pith(a_t \given s_t) \sum_{\delay=0}^{\infty} (\gamma \lambda)^{\delay}\dv_{t+\delay}},
\label{eq:pg-gae1}
\end{align}
where equality holds when $\lambda=1$.

\section{Interpretation as Reward Shaping}
\label{sec:shaping}

In this section, we discuss how one can interpret $\lambda$ as an extra discount factor applied after performing a reward shaping transformation on the MDP.
We also introduce the notion of a response function to help understand the bias introduced by $\gamma$ and $\lambda$.

Reward shaping \citep{ng1999policy} refers to the following transformation of the reward function of an MDP:
let $\Phi: \mathcal{S} \rightarrow \Real$ be an arbitrary scalar-valued function on state space, and define the transformed reward function $\tilr$ by
\begin{align}
\tilr(s,a,s') = r(s,a,s') + \gamma \Phi(s') - \Phi(s), \label{eq:reshaping}
\end{align}
which in turn defines a transformed MDP.
This transformation leaves the discounted advantage function $\Apigam$ unchanged for any policy $\pi$.
To see this, consider the discounted sum of rewards of a trajectory starting with state $s_t$:
\begin{align}
\sum_{\delay=0}^{\infty} \gamma^{\delay}\tilr(s_{t+\delay},a_t,s_{t+\delay+1}) &= \sum_{\delay=0}^{\infty} \gamma^{\delay}r(s_{t+\delay},a_{t+\delay},s_{t+\delay+1}) - \Phi(s_{t}).
\label{eq:dsshaped}
\end{align}
Letting $\tilde{Q}^{\pi,\gamma},\tilde{V}^{\pi,\gamma},\tilde{A}^{\pi,\gamma}$ be the value and advantage functions of the transformed MDP, one obtains from the definitions of these quantities that
\begin{align}
\tilde{Q}^{\pi,\gamma}(s, a) &= \Qpigam(s,a) - \Phi(s)\\
\tilde{V}^{\pi,\gamma}(s, a) &= \Vpigam(s) - \Phi(s)\\
\tilde{A}^{\pi,\gamma}(s, a) &= (\Qpigam(s,a) - \Phi(s)) - (\Vpigam(s) - \Phi(s)) = \Apigam(s,a).
\end{align}
Note that if $\Phi$ happens to be the state-value function $\Vpigam$ from the original MDP, then the transformed MDP has the interesting property that $\tilde{V}^{\pi,\gamma}(s)$ is zero at every state.

Note that \citep{ng1999policy} showed that the reward shaping transformation leaves the policy gradient and optimal policy unchanged when our objective is to maximize the discounted sum of rewards $\sum_{t=0}^{\infty} \gamma^t r(s_t, a_t, s_{t+1})$.
In contrast, this paper is concerned with maximizing the undiscounted sum of rewards, where the discount $\gamma$ is used as a variance-reduction parameter.

Having reviewed the idea of reward shaping, let us consider how we could use it to get a policy gradient estimate.
The most natural approach is to construct policy gradient estimators that use discounted sums of shaped rewards $\tilr$.
However, \Cref{eq:dsshaped} shows that we obtain the discounted sum of the original MDP's rewards $r$ minus a baseline term.
Next, let's consider using a ``steeper'' discount $\gamma \lambda$, where $0 \le \lambda \le 1$.
It's easy to see that the shaped reward $\tilr$ equals the Bellman residual term $\dv$, introduced in \Cref{sec:advest}, where we set $\Phi=V$.
Letting $\Phi=V$, we see that
\begin{align}
\sum_{\delay=0}^{\infty} (\gamma \lambda)^{\delay}\tilr(s_{t+\delay},a_t,s_{t+\delay+1}) &= \sum_{\delay=0}^{\infty} (\gamma \lambda)^{\delay} \dv_{t+\delay}
= \hatalam_t.
\end{align}
Hence, by considering the $\gamma\lambda$-discounted sum of shaped rewards, we exactly obtain the generalized advantage estimators from \Cref{sec:advest}.
As shown previously, $\lambda=1$ gives an unbiased estimate of $\bgrad$, whereas $\lambda<1$ gives a biased estimate.

To further analyze the effect of this shaping transformation and parameters $\gamma$ and $\lambda$, it will be useful to introduce the notion of a response function $\resp$, which we define as follows:
\begin{align}
\resp(\delay; s_t, a_t) = \Ea{r_{t+\delay} \given s_t, a_t}  - \Ea{r_{t+\delay} \given s_t}.
\end{align}
Note that $A^{\pi,\gamma}(s,a) = \sum_{\delay=0}^{\infty} \gamma^{\delay} \resp(\delay; s, a)$, hence the response function decomposes the advantage function across timesteps.
The response function lets us quantify the temporal credit assignment problem: long range dependencies between actions and rewards correspond to nonzero values of the response function for $\delay \gg 0$.

Next, let us revisit the discount factor $\gamma$ and the approximation we are making by using $\Apigam$ rather than $A^{\pi,1}$.
The discounted policy gradient estimator from \Cref{eq:pg-adv-biased} has a sum of terms of the form
\begin{align}
\gradth \log \pith(a_t \given s_t) A^{\pi,\gamma}(s_t, a_t)
&= \gradth \log \pith(a_t \given s_t) \sum_{\delay=0}^{\infty} \gamma^{\delay} \chi(\delay; s_t,a_t) .
\end{align}
Using a discount $\gamma < 1$ corresponds to dropping the terms with $\delay \gg 1/(1-\gamma)$.
Thus, the error introduced by this approximation will be small if $\resp$ rapidly decays as $\delay$ increases, i.e., if the effect of an action on rewards is ``forgotten'' after $\approx 1/(1-\gamma)$ timesteps.

If the reward function $\tilr$ were obtained using $\Phi=\Vpigam$, we would have $\Ea{\tilr_{t+\delay} \given s_t,a_t} = \Ea{\tilr_{t+\delay} \given s_t} = 0$ for $\delay > 0$, i.e., the response function would only be nonzero at $\delay=0$.
Therefore, this shaping transformation would turn temporally extended response into an immediate response.
Given that $\Vpigam$ completely reduces the temporal spread of the response function, we can hope that a good approximation $V \approx \Vpigam$ partially reduces it.
This observation suggests an interpretation of \Cref{eq:gaelam1}: reshape the rewards using $V$ to shrink the temporal extent of the response function, and then introduce a ``steeper'' discount $\gamma \lambda$ to cut off the noise arising from long delays, i.e., ignore terms $\gradth \log \pith(a_t \given s_t) \dv_{t+\delay}$ where $\delay \gg 1/(1-\gamma \lambda)$.

\section{Value Function Estimation}

A variety of different methods can be used to estimate the value function (see, e.g., \cite{bertsekas1995dynamic}).
When using a nonlinear function approximator to represent the value function, the simplest approach is to solve a nonlinear regression problem:
\begin{align}
\minimize_{\phi} \sum_{n=1}^N \norm{ V_{\phi}(s_n) - \Vhat_n }^2,
\label{eq:nonlinearreg}
\end{align}
where $\Vhat_t = \sum_{l=0}^{\infty} \gamma^l r_{t+l}$ is the discounted sum of rewards,
and $n$ indexes over all timesteps in a batch of trajectories.
This is sometimes called the Monte Carlo or \tdone{} approach for estimating the value function \citep{sutton1998introduction}.\footnote{Another natural choice is to compute target values with an estimator based on the \tdlam{} backup \citep{bertsekas1995dynamic,sutton1998introduction}, mirroring the expression we use for policy gradient estimation: $\hat{V^{\lambda}_t} =  V_{\phold}(s_n) + \sum_{\delay=0}^{\infty} (\gamma \lambda)^{\delay} \delta_{t+\delay}$.
While we experimented with this choice, we did not notice a difference in performance from the $\lambda=1$ estimator in \Cref{eq:nonlinearreg}.}

For the experiments in this work, we used a trust region method to
optimize the value function in each iteration of a batch optimization procedure.
The trust region helps us to avoid overfitting to the most recent batch of data.
To formulate the trust region problem, we first compute $\sigma^2 = \frac{1}{N} \sum_{n=1}^N \norm{ V_{\phold}(s_n) - \Vhat_n }^2$, where $\phold$ is the parameter vector before optimization.
Then we solve the following constrained optimization problem:
\begin{flalign}
&\minimize_{\phi} \quad \sum_{n=1}^N \norm{ V_{\phi}(s_n) - \Vhat_n }^2
\nonumber \\
&\text{subject to} \quad \frac{1}{N} \sum_{n=1}^N \frac{\norm{ V_{\phi}(s_n) - V_{\phold}(s_n) }^2}{2\sigma^2} \le \epsilon.
\label{eq:vfprob}
\end{flalign}
This constraint is equivalent to constraining the average KL divergence between the previous value function and the new value function to be smaller than $\epsilon$, where the value function is taken to parameterize a conditional  Gaussian distribution with mean $V_{\phi}(s)$ and variance $\sigma^2$.

We compute an approximate solution to the trust region problem using the conjugate gradient algorithm \citep{wright1999numerical}.
Specifically, we are solving the quadratic program
\begin{align}
&\minimize_{\phi} \quad g^T (\phi - \phold) \nonumber \\
&\text{subject to} \quad \frac{1}{N} \sum_{n=1}^N (\phi-\phold)^T H (\phi-\phold) \le \epsilon.
\label{eq:vftr}
\end{align}
where $g$ is the gradient of the objective, and $H=\frac{1}{N}\sum_n j_n j_n^T$, where $j_n = \grad_{\phi} V_{\phi}(s_n)$.
Note that $H$ is the ``Gauss-Newton'' approximation of the Hessian of the objective, and it is (up to a $\sigma^2$ factor) the Fisher information matrix when interpreting the value function as a conditional probability distribution.
Using matrix-vector products $v \rightarrow Hv$ to implement the conjugate gradient algorithm, we compute a step direction $s \approx -H^{-1}g$. Then we rescale $s\rightarrow \alpha s$ such that $\frac{1}{2} (\alpha s)^T H (\alpha s) = \epsilon$ and take $\phi=\phold + \alpha s$.
This procedure is analogous to the procedure we use for updating the policy, which is described further in \Cref{sec:experiments} and based on \cite{schulman2015trust}.

\section{Experiments} \label{sec:experiments}

We designed a set of experiments to investigate the following questions:
\begin{enumerate}[leftmargin=*]
\item What is the empirical effect of varying $\lambda \in [0,1]$ and $\gamma \in [0,1]$ when optimizing episodic total reward using generalized advantage estimation?
\item Can generalized advantage estimation, along with trust region algorithms for policy and value function optimization, be used to optimize large neural network policies for challenging control problems?
\end{enumerate}

\subsection{Policy Optimization Algorithm}

While generalized advantage estimation can be used along with a variety of different policy gradient methods, for these experiments, we performed the policy updates using trust region policy optimization  (TRPO) \citep{schulman2015trust}.
TRPO updates the policy by approximately solving the following constrained optimization problem each iteration:
\newcommand{\pithold}{\pi_{\thold}}
\begin{align}
&\minimize_{\theta} L_{\thold}(\theta) \nonumber\\
& \text{subject to } \ \meankl{\thold}(\pithold,\pith) \le \epsilon \nonumber\\
& \text{where } L_{\thold}(\theta) = \frac{1}{N}\sum_{n=1}^{N}
\frac{\pith(a_n \given s_n)}{\pithold(a_n\given s_n)}\hata_n\nonumber\\
& \text{\ \ \ \ \ \ \ \ \ \ } \meankl{\thold}(\pithold,\pith) = \frac{1}{N}\sum_{n=1}^{N} \kl{\pithold(\cdot \given s_n)}{\pith(\cdot  \given s_n)}
\label{eq:trpo}
\end{align}
As described in \citep{schulman2015trust}, we approximately solve this problem by linearizing the objective and quadraticizing the constraint, which yields a step in the direction $\theta - \thold \propto -F^{-1}g$, where $F$ is the average Fisher information matrix, and $g$ is a policy gradient estimate.
This policy update yields the same step direction as the natural policy gradient \citep{kakade2001natural} and natural actor-critic \citep{peters2008natural}, however it uses a different stepsize determination scheme and numerical procedure for computing the step.

Since prior work \citep{schulman2015trust} compared TRPO to a variety of different policy optimization algorithms, we will not repeat these comparisons; rather, we will focus on varying the $\gamma,\lambda$ parameters of policy gradient estimator while keeping the underlying algorithm fixed.

For completeness, the whole algorithm for iteratively updating policy and value function is given below:
\newcommand{\pithi}{\pi_{\theta_i}}
\begin{algorithm}
\begin{algorithmic}
\State Initialize policy parameter $\theta_0$ and value function parameter $\phi_0$.
\For{$i$ = 0, 1, 2, \dots}
\State Simulate current policy $\pithi$ until $N$ timesteps are obtained.
\State Compute $\dv_t$ at all timesteps $t \in \lrbrace{1, 2, \dots,
N}$, using $V=V_{\phi_{i}}$.
\State Compute $\hata_t = \sum_{\delay=0}^{\infty} (\gamma \lambda)^{\delay} \dv_{t+\delay}$ at all timesteps.
\State Compute $\theta_{i+1}$ with TRPO update, \Cref{eq:trpo}.
\State Compute $\phi_{i+1}$ with \Cref{eq:vftr}. 
\EndFor
\label{alg}
\end{algorithmic}
\end{algorithm}

Note that the policy update $\theta_i \rightarrow \theta_{i+1}$ is performed using the value function $V_{\phi_i}$ for advantage estimation, not $V_{\phi_{i+1}}$.
Additional bias would have been introduced if we updated the value function first.
To see this, consider the extreme case where we overfit the value function, and the Bellman residual $r_t + \gamma V(s_{t+1}) - V(s_t)$ becomes zero at all timesteps---the policy gradient estimate would be zero.
\subsection{Experimental Setup}

\begin{figure}[htp]
\centering
\includegraphics[width=0.4\columnwidth]{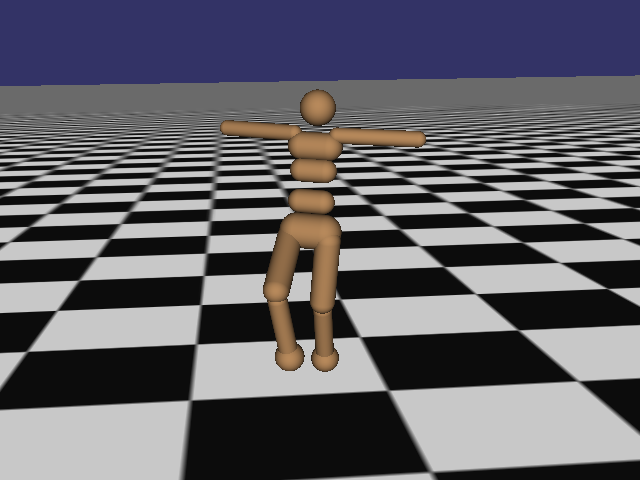}
\hspace{0cm}
\includegraphics[width=.4\columnwidth]{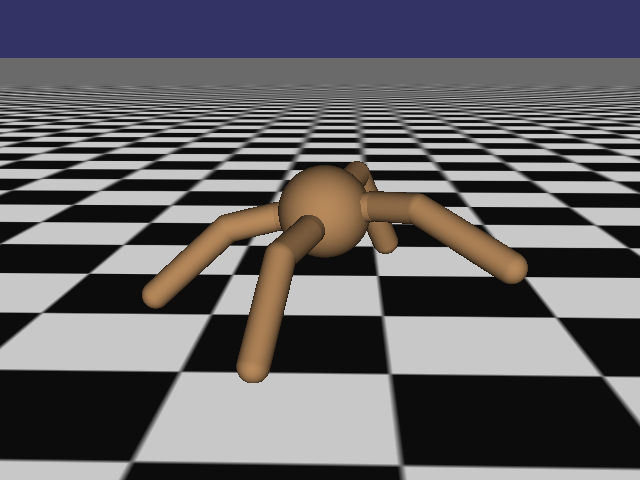}
\\
\includegraphics[width=0.4\columnwidth]{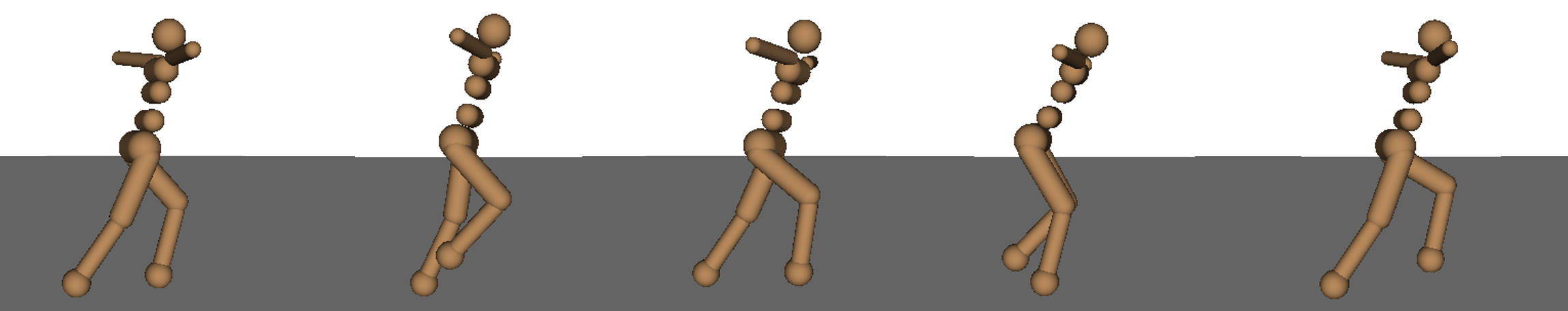}
\hspace{0cm}
\includegraphics[width=0.4\columnwidth]{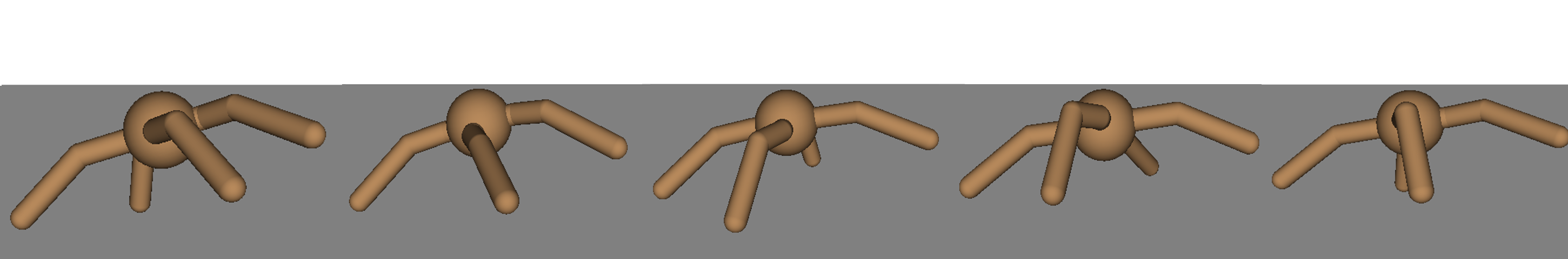}
\caption{Top figures: robot models used for 3D locomotion. Bottom figures: a sequence of frames from the learned gaits. Videos are available at \url{https://sites.google.com/site/gaepapersupp}.}
\label{fig:humanoidmodel}
\end{figure}

We evaluated our approach on the classic cart-pole balancing problem, as well as several challenging 3D locomotion tasks: (1) bipedal locomotion; (2) quadrupedal locomotion; (3) dynamically standing up, for the biped, which starts off laying on its back.
The models are shown in \Cref{fig:humanoidmodel}.

\subsubsection{Architecture}

We used the same neural network architecture for all of the 3D robot tasks, which was a feedforward network with three hidden layers, with $100$, $50$ and $25$ tanh units respectively.
The same architecture was used for the policy and value function.
The final output layer had linear activation.
The value function estimator used the same architecture, but with only one scalar output.
For the simpler cart-pole task, we used a linear policy, and a neural network with one 20-unit hidden layer as the value function.

\subsubsection{Task details}

For the cart-pole balancing task, we collected 20 trajectories per batch, with a maximum length of 1000 timesteps, using the physical parameters from \citet{barto1983neuronlike}.

The simulated robot tasks were simulated using the MuJoCo physics engine \citep{todorov2012mujoco}.
The humanoid model has 33 state dimensions and 10 actuated degrees of freedom, while the quadruped model has 29 state dimensions and 8 actuated degrees of freedom. The initial state for these tasks consisted of a uniform distribution centered on a reference configuration.
We used 50000 timesteps per batch for bipedal locomotion, and 200000 timesteps per batch for quadrupedal locomotion and bipedal standing.
Each episode was terminated after $2000$ timesteps if the robot had not reached a terminal state beforehand.
The timestep was $0.01$ seconds.

The reward functions are provided in the table below.
\newcommand{\fimp}{f_{\mathrm{impact}}}
\begin{equation*} 
\begin{tabular}{cc}
Task & Reward \\
\hline
3D biped locomotion & $v_{\mathrm{fwd}} - 10^{-5} \norm{u}^2 - 10^{-5} \norm{\fimp}^2 + 0.2$\\
Quadruped locomotion & $v_{\mathrm{fwd}} - 10^{-6} \norm{u}^2 - 10^{-3}\norm{\fimp}^2 + 0.05$\\
Biped getting up & $-(h_{\rm head} - 1.5)^2 - 10^{-5} \norm{u}^2$\\
\end{tabular}
\end{equation*}

Here, $v_{\mathrm{fwd}} \defeq \text{forward velocity}$, $u \defeq \text{vector of joint torques}$, $\fimp \defeq \text{impact forces}$, $h_{\rm head}\defeq \text{height of the head}$.

In the locomotion tasks, the episode is terminated if the center of mass of the actor falls below a predefined height: $\unit[.8]{m}$ for the biped, and $\unit[.2]{m}$ for the quadruped.
The constant offset in the reward function encourages longer episodes; otherwise the quadratic reward terms might lead lead to a policy that ends the episodes as quickly as possible.

\subsection{Experimental Results}
All results are presented in terms of the cost, which is defined as negative reward and is minimized. Videos of the learned policies are available at \url{https://sites.google.com/site/gaepapersupp}.
In plots, ``No VF'' means that we used a time-dependent baseline that did not depend on the state, rather than an estimate of the state value function.
The time-dependent baseline was computed by averaging the return at each timestep over the trajectories in the batch.

\subsubsection{Cart-pole}

\begin{figure}[h!]
\centering
\includegraphics[width=0.45\textwidth]{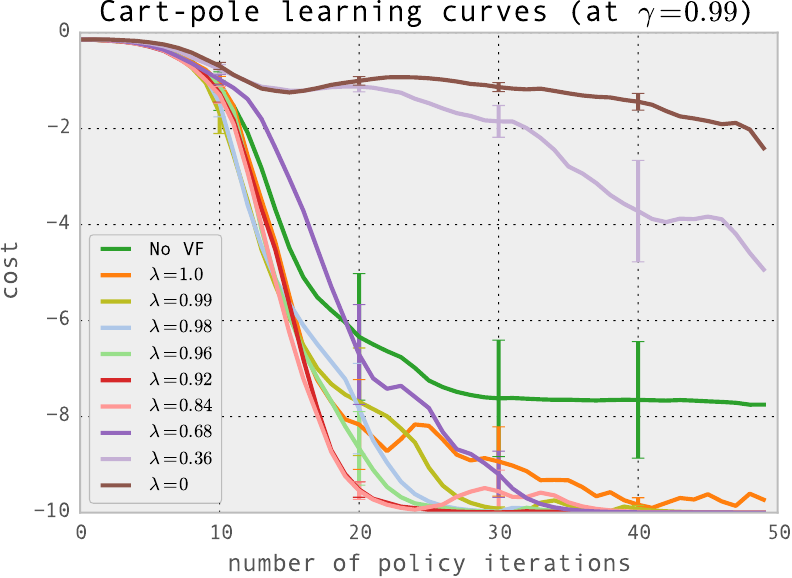}
\hspace{.5cm}
\includegraphics[width=0.45\textwidth]{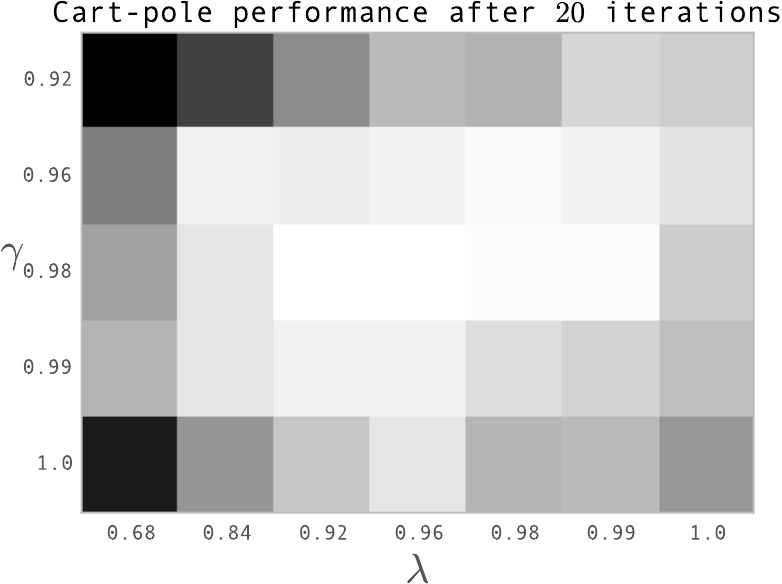}
\caption{
Left: learning curves for cart-pole task, using generalized advantage estimation with varying values of $\lambda$ at $\gamma=0.99$.
The fastest policy improvement is obtain by intermediate values of $\lambda$ in the range $[0.92, 0.98]$.
Right: performance after 20 iterations of policy optimization, as $\gamma$ and $\lambda$ are varied. White means higher reward. The best results are obtained at intermediate values of both.
}
\label{fig:cartpoley}
\end{figure}

The results are averaged across $21$ experiments with different random seeds.
Results are shown in \Cref{fig:cartpoley}, and indicate that the best results are obtained at intermediate values of the parameters: $\gamma \in [0.96, 0.99]$ and $\lambda \in [0.92, 0.99]$.

\subsubsection{3D bipedal locomotion}

Each trial took about 2 hours to run on a 16-core
machine, where the simulation rollouts were parallelized, as were the function, gradient, and matrix-vector-product evaluations used when optimizing the policy and value function.
Here, the results are averaged across $9$ trials with different random seeds.
The best performance is again obtained using intermediate values of $\gamma \in [0.99, 0.995], \lambda \in [0.96, 0.99]$.
The result after 1000 iterations is a fast, smooth, and stable gait that is effectively completely stable.
We can compute how much ``real time'' was used for this learning process: $\unitfrac[0.01]{seconds}{timestep} \times \unitfrac[50000]{timesteps}{batch} \times \unit[1000]{batches} / \unitfrac[3600 \cdot 24]{seconds}{day}=\unit[5.8]{days}$.
Hence, it is plausible that this algorithm could be run on a real robot, or multiple real robots learning in parallel, if there were a way to reset the state of the robot and ensure that it doesn't damage itself.

\begin{figure}
\centering
\includegraphics[width=.48\textwidth]{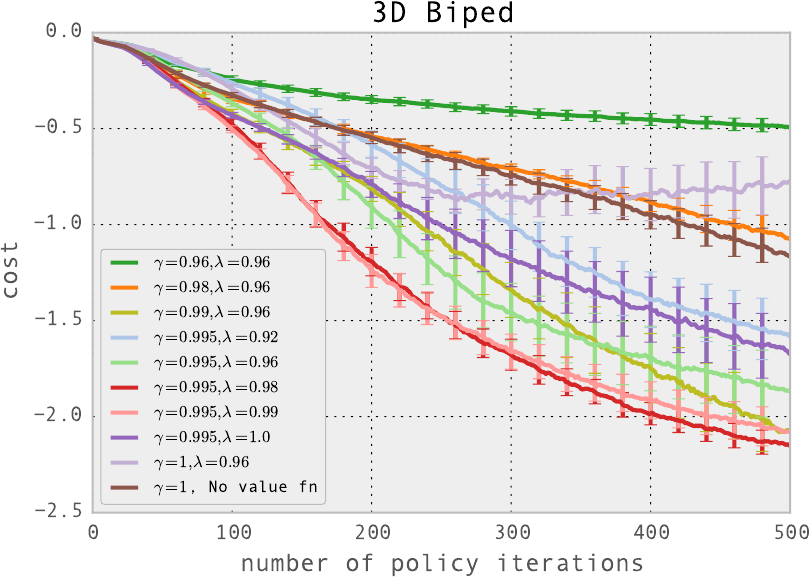}
\includegraphics[width=.48\textwidth]{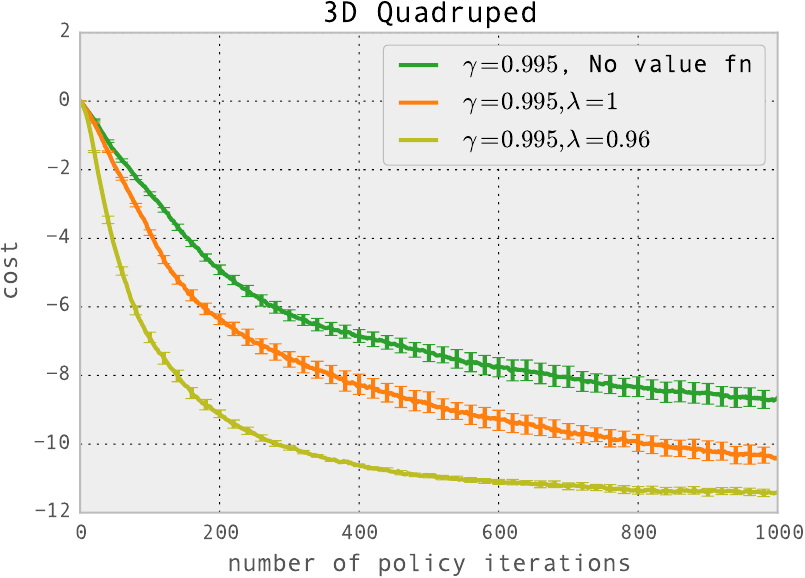}
\label{fig:humanoidlc}
\caption{Left: Learning curves for 3D bipedal locomotion, averaged across nine runs of the algorithm. Right: learning curves for 3D quadrupedal locomotion, averaged across five runs.}
\end{figure}

\subsubsection{Other 3D robot tasks}
The other two motor behaviors considered are quadrupedal locomotion and getting up off the ground for the 3D biped.
Again, we performed 5 trials per experimental condition, with different random seeds (and initializations). The experiments took about 4 hours per trial on a 32-core machine.
We performed a more limited comparison on these domains (due to the substantial computational resources required to run these experiments), fixing $\gamma=0.995$ but varying $\lambda = \lrbrace{0,0.96}$, as well as an experimental condition with no value function.
For quadrupedal locomotion, the best results are obtained using a value function with $\lambda=0.96$ \Cref{fig:humanoidlc}.
For 3D standing, the value function always helped, but the results are roughly the same for $\lambda=0.96$ and $\lambda=1$.

\begin{figure}[!h]
\centering
\includegraphics[width=.45\textwidth]{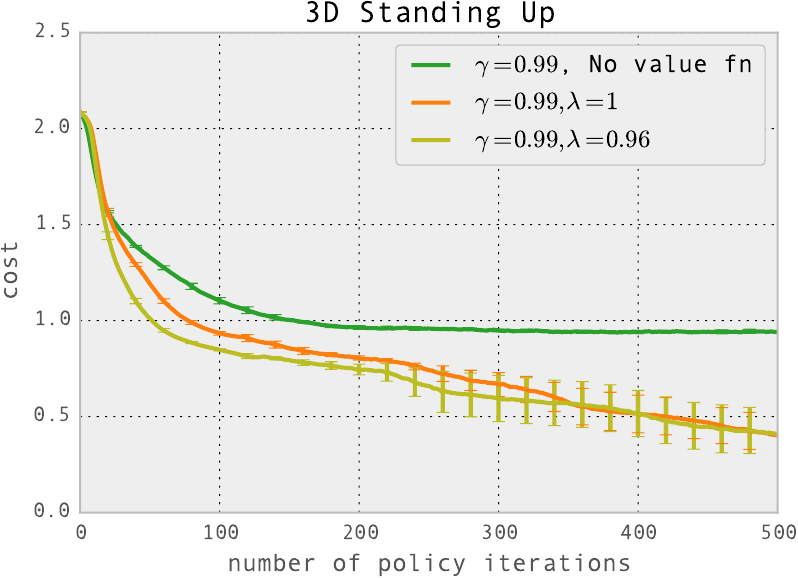}
\includegraphics[width=.45\textwidth]{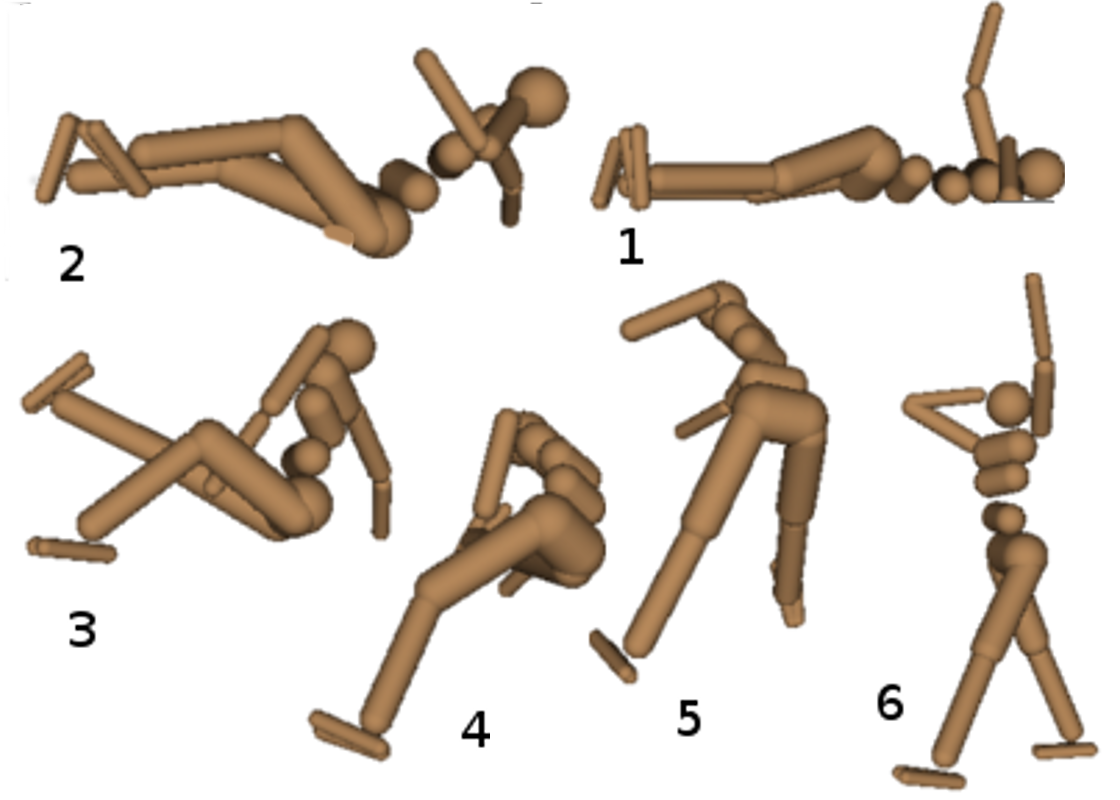}
\caption{(a) Learning curve from quadrupedal walking, (b) learning curve for 3D standing up, (c) clips from 3D standing up.}
\end{figure}

\section{Discussion}

Policy gradient methods provide a way to reduce reinforcement learning to stochastic gradient descent, by providing unbiased gradient estimates.
However, so far their success at solving difficult control problems has been limited, largely due to their high sample complexity.
We have argued that the key to variance reduction is to obtain good estimates of the advantage function.

We have provided an intuitive but informal analysis of the problem of
advantage function estimation, and justified the generalized advantage
estimator, which has two parameters $\gamma,\lambda$ which adjust the bias-variance tradeoff.
We described how to combine this idea with trust region policy optimization and a trust region algorithm that optimizes a value function, both represented by neural networks.
Combining these techniques, we are able to learn to solve difficult control tasks that have previously been out of reach for generic reinforcement learning methods.

Our main experimental validation of generalized advantage estimation is in the domain of simulated robotic locomotion. As shown in our experiments, choosing an appropriate intermediate value of $\lambda$ in the range $[0.9, 0.99]$ usually results in the best performance.
A possible topic for future work is how to adjust the estimator parameters $\gamma,\lambda$ in an adaptive or automatic way.

One question that merits future investigation is the relationship between value function estimation error and policy gradient estimation error.
If this relationship were known, we could choose an error metric for value function fitting that is well-matched to the quantity of interest, which is typically the accuracy of the policy gradient estimation.
Some candidates for such an error metric might include the Bellman error or projected Bellman error, as described in \cite{bhatnagar2009convergent}.

Another enticing possibility is to use a shared function approximation architecture for the policy and the value function, while optimizing the policy using generalized advantage estimation.
While formulating this problem in a way that is suitable for numerical optimization and provides convergence guarantees remains an open question, such an approach could allow the value function and policy representations to share useful features of the input, resulting in even faster learning.

In concurrent work, researchers have been developing policy gradient methods that involve differentiation with respect to the continuous-valued action \citep{lillicrap2015continuous,heess2015learning}.
While we found empirically that the one-step return ($\lambda=0$) leads to excessive bias and poor performance, these papers show that such methods can work when tuned appropriately.
However, note that those papers consider control problems with substantially lower-dimensional state and action spaces than the ones considered here.
A comparison between both classes of approach would be useful for future work.

\section*{Acknowledgements}
We thank Emo Todorov for providing the simulator as well as insightful discussions, and we thank Greg Wayne, Yuval Tassa, Dave Silver, Carlos Florensa Campo, and Greg Brockman for insightful discussions.
This research was funded in part by the Office of Naval Research through a Young Investigator Award and under grant number N00014-11-1-0688,
DARPA through a Young Faculty Award, by the Army Research Office through the MAST program.

\appendix

\section{Frequently Asked Questions}

\newcommand{\gradthi}{\grad_{\theta_i}}
\newcommand{\gradthj}{\grad_{\theta_j}}

\subsection{What's the Relationship with Compatible Features?}\label{sec:cf}

Compatible features are often mentioned in relation to policy gradient algorithms that make use of a value function, and the idea was proposed in the paper \textit{On Actor-Critic Methods} by \citet{konda2003onactor}.
These authors pointed out that due to the limited representation power of the policy, the policy gradient only depends on a certain subspace of the space of advantage functions.
This subspace is spanned by the compatible features $\gradthi \log \pith(a_t \given s_t)$, where $i \in \lrbrace{1,2,\dots,\dim \theta}$.
This  theory of compatible features provides no guidance on how to exploit the temporal structure of the problem to obtain better estimates of the advantage function, making it mostly orthogonal to the ideas in this paper.

The idea of compatible features motivates an elegant method for computing the natural policy gradient \citep{kakade2001natural,peters2008natural}.
Given an empirical estimate of the advantage function $\hata_t$ at each timestep, we can project it onto the subspace of compatible features by solving the following least squares problem:
\begin{align}
\minimize_{\br} \sum_t \norm{\br \cdot \gradth \log \pith(a_t \given s_t) - \hata_t}^2.
\end{align}
If $\hata$ is \just{}, the least squares solution is the natural policy gradient \citep{kakade2001natural}.
Note that any estimator of the advantage function can be substituted into this formula, including the ones we derive in this paper.
For our experiments, we also compute natural policy gradient steps, but we use the more computationally efficient numerical procedure from \cite{schulman2015trust}, as discussed in \Cref{sec:experiments}.

\subsection{Why Don't You Just Use a $Q$-Function?}
Previous actor critic methods, e.g. in \cite{konda2003onactor}, use a $Q$-function to obtain potentially low-variance policy gradient estimates.
Recent papers, including \cite{heess2015learning,lillicrap2015continuous}, have shown that a neural network $Q$-function approximator can used effectively in a policy gradient method.
However, there are several advantages to using a state-value function in the manner of this paper.
First, the state-value function has a lower-dimensional input and is thus easier to learn than a state-action value function.
Second, the method of this paper allows us to smoothly interpolate between the high-bias estimator ($\lambda=0$) and the low-bias estimator ($\lambda=1$).
On the other hand, using a parameterized $Q$-function only allows us to use a high-bias estimator.
We have found that the bias is prohibitively large when using a one-step estimate of the returns, i.e., the $\lambda=0$ estimator, $\hata_t = \dv_t = r_t + \gamma V(s_{t+1}) - V(s_t)$.
We expect that similar difficulty would be encountered when using an advantage estimator involving a parameterized $Q$-function, $\hata_t = Q(s,a) - V(s)$.
There is an interesting space of possible algorithms that would use a parameterized $Q$-function and attempt to reduce bias, however, an exploration of these possibilities is beyond the scope of this work.

\section{Proofs}\label{sec:proofs}
\textbf{Proof of Proposition \hyperlink{justprop}{1}}:
First we can split the expectation into terms involving $Q$ and $b$,
\begin{align}
&\Eb{s_{0:\infty},a_{0:\infty}}{\gradth \log \pith(a_t \given s_t) (\qofall-\bofpast)}  \nonumber\\
&\ \ \ \ = \Eb{s_{0:\infty},a_{0:\infty}}{\gradth \log \pith(a_t \given s_t) (\qofall)}\nonumber \\
&\ \ \ \ \ \ - \Eb{s_{0:\infty},a_{0:\infty}}{\gradth \log \pith(a_t \given s_t) (\bofpast)}
\end{align}
We'll consider the terms with $Q$ and $b$ in turn.
\begin{align*}
&\!\!\!\! \Eb{s_{0:\infty},a_{0:\infty}}{\gradth \log \pith(a_t \given s_t) \qofall} \\
&=\Eb{s_{0:t},a_{0:t}}{\Eb{s_{t+1:\infty},a_{t+1:\infty}}{\gradth \log \pith(a_t \given s_t) \qofall}}\\
&=\Eb{s_{0:t},a_{0:t}}{ \gradth \log \pith(a_t \given s_t) \Eb{s_{t+1:\infty},a_{t+1:\infty}}{\qofall}}\\
&=\Eb{s_{0:t},a_{0:t-1}}{ \gradth \log \pith(a_t \given s_t) \Api(s_t, a_t)}
\end{align*}
Next,
\begin{align*}
&\!\!\!\!\Eb{s_{0:\infty},a_{0:\infty}}{\gradth \log \pith(a_t \given s_t) \bofpast} \\
&=\Eb{s_{0:t},a_{0:t-1}}{\Eb{s_{t+1:\infty},a_{t:\infty}}{\gradth \log \pith(a_t \given s_t) \bofpast}}\\
&=\Eb{s_{0:t},a_{0:t-1}}{ \Eb{s_{t+1:\infty},a_{t:\infty}}{\gradth \log \pith(a_t \given s_t)}\bofpast}\\
&=\Eb{s_{0:t},a_{0:t-1}}{ 0 \cdot \bofpast}\\
&=0.
\end{align*}

{\small
\bibliographystyle{iclr2016_conference}
\bibliography{vfuncs}
}

\end{document}